\documentclass{article}

\usepackage{microtype}
\usepackage{graphicx}
\usepackage{subfigure}
\usepackage{booktabs} 
\usepackage{amsfonts}
\usepackage{mathptmx}
\usepackage{amsmath}
\usepackage{hyperref}

\usepackage[accepted]{icml2019}

\icmltitlerunning{Perceptual Values from Observation}

\begin{document}

\twocolumn[
\icmltitle{Perceptual Values from Observation}



\begin{icmlauthorlist}
\icmlauthor{Ashley D. Edwards}{to}
\icmlauthor{Charles L. Isbell}{to}
\end{icmlauthorlist}

\icmlaffiliation{to}{Georgia Institute of Technology, Atlanta, GA, USA}

\icmlcorrespondingauthor{Ashley D. Edwards}{aedwards8@gatech.edu}

\icmlkeywords{Machine Learning, ICML}

\vskip 0.3in
]



\printAffiliationsAndNotice{}  

\begin{abstract}
Imitation by observation is an approach for learning from expert demonstrations that lack action information, such as videos. Recent approaches to this problem can be placed into two broad categories: training dynamics models that aim to predict the actions taken between states, and learning rewards or features for computing them for Reinforcement Learning (RL). In this paper, we introduce a novel approach that learns values, rather than rewards, directly from observations. We show that by using values, we can significantly speed up RL by removing the need to bootstrap action-values, as compared to sparse-reward specifications. 
\end{abstract}

\section{Introduction}
When people solve tasks, there is often a clear ordering to which steps are more preferable than others. For example, when building a piece of furniture, we may assume that it is better to have the pieces outside of the box than in, and a leg screwed into its base than laying on the floor. The final steps of a problem are often more desirable than earlier ones, assuming the task is being completed optimally, because they indicate that we have fewer steps left to go. Put in other terms, these later steps are typically more~\emph{valuable} than those seen in the beginning of the problem.

In this paper, we use this insight to compute values from expert observations without access to the underlying actions or rewards. Because task goals are often achieved at or near the end of a demonstration trajectory, it is likely that later states should have more value than early ones. Hence, given expert state trajectories, we compute the expected value of each state by assuming that the reward at the last state in the trajectory is 1 and 0 everywhere else, and then backing up values to the start of the trace by utilizing knowledge of the length of the trajectory in a self-supervised manner. We show how these values can be used to learn action-values for reinforcement learning more efficiently than training from sparse rewards.

We formally introduce our approach, Perceptual Values from Observation (PVO), which aims to learn values directly from expert observations. We show that this approach learns meaningful values that increase as the goal nears, and demonstrate that these values can be used to train a reinforcement learning agent. We demonstrate the learned values in a maze environment~\cite{mazelab},  liquid pouring task~\cite{sermanet2016unsupervised}, and a task for picking up objects~\cite{goyal2017something}, and show that PVO can be used to train RL agents in OpenAI's CoinRun environment~\cite{cobbe2018quantifying}. 

\section{Related work}
There has recently been a large amount of interest in learning behaviors from expert state observations. This mechanism introduces several opportunities to obtain training examples for agents; there is a wealth of pre-existing videos that consist of humans and other entities---such as animated characters and animals---performing tasks that we might like an agent to learn. Learning in this manner becomes more difficult however because the underlying actions and rewards are unknown. In order to make use of the abundance of video data available on the web, we should consider how we can learn goals and values without access to this information. 

Given a set of single-goal observations, one recent approach is to train a classifier to predict if a state is a goal or not and then use this discriminator as a reward signal~\cite{xie2018few, singh2019}. However, goal-prediction is essentially a sparse-reward problem and thus may not shape behavior. As such, while single-goal representations require little demonstration data, they may require more environment interactions to train reinforcement learning agents than methods that provide more guidance. In general, we should expect a trade-off between the amount of experience we need to provide an agent and the amount of time it will take the agent to learn. 

To that point, we can train models using already existing videos or other forms of observation. This work focuses on learning to imitate from such sequences. One approach to this problem is to learn or use pre-existing features for computing rewards~\cite{edwards2016perceptual, liu2017imitation, sermanet2017time, aytar2018playing, yu2019unsupervised}. Such approaches likely offer a better shaped reward than goal-prediction based rewards because they are based on the distance to the goal. Another approach is to learn rewards directly~\cite{sermanet2016unsupervised, edwards2017cross} or in an adversarial manner~\cite{torabi2018generative}. Finally, we can avoid learning rewards at all by learning dynamics that aim to infer the actions taken in the state sequences~\cite{pathak2018zero, torabi2018behavioral, edwards2018imitating}. However, learning dynamics can often be difficult and may require a large amount of demonstration or environmental data.

This paper introduces another mechanism for learning from state observations. In particular, we are interested in learning values because they allow us to bypass engineering reward functions that may be susceptible to locally sub-optimal solutions. As we will show, by using values we can additionally remove the bootstrapped component of training reinforcement learning.

\section{Formalities}
We are interested in solving problems specified through a Markov Decision Process, where we do not have access to the transition function or environment rewards, and the states consist of visual inputs. We are given a set of expert state observations $D$ where we assume we also do not have access to the underlying expert actions or rewards. 

\section{Approach}
\begin{figure}[htb]
    \centering
    \includegraphics[width=.8\linewidth]{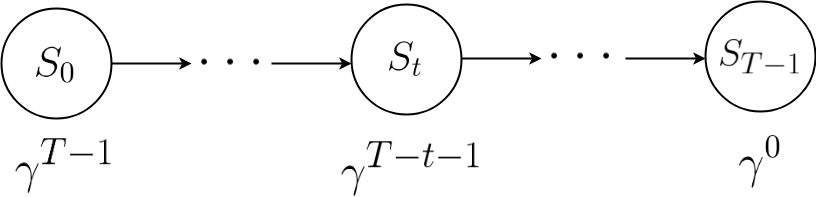}
    \caption{Value assignment for a length $T$ trajectory sampled from expert demonstrations. We can train these values in a self-supervised manner by utilizing the number of steps a state is from the end of the trajectory.}
    \label{fig:value_assignment}
\end{figure}

Given a trajectory of expert observations $\{s^*_0 \dots s^*_n\} \in D$, PVO aims to learn a value function $V_\theta$ that makes an approximation of the expert value function. As we noted, we are not given the underlying reward function with these demonstrations. Rather, we enforce a  surrogate reward based on a simple assumption that tasks obtained from expert observations can be specified through a sparse reward of 1 at the end of the trajectory and 0 elsewhere.

This hypothesis comes from the observation that the goal will often occur at the end of the trajectory, especially in goal-directed tasks. However, we enforce this reward function even if a trajectory does not actually end at the goal. Using this assumption, we may backtrack values from the end of a trajectory to the start without knowing the actions taken.  We then use this value function to learn values of novel states and to learn action-values for RL. 

\subsection{Step 1: Learning values from observation}
 \begin{algorithm}[h]
\begin{algorithmic}[1]
	\Function{PVO}{$D$}
    	\For{$k \gets 0 \ldots \text{\#\it{Epochs}}$} 
        	\State $\langle s^*_0$, $s^*_1$, \ldots, $s^*_T \rangle \sim D$ 
        	\State $t \sim uniform(0, T)$
        	\State $y_k = \gamma^{T-t-1}$ 
        	\State $\theta_{k+1} \gets \theta_k - \alpha\nabla_\theta \Vert y_k - V_\theta(s_{t})\Vert^2$ 
        \EndFor
    \caption{Perceptual Values from Observation}
    \EndFunction
\label{algo:pvo}
\end{algorithmic}
\end{algorithm}

The first step of this approach aims to obtain values from expert observations. Given a length $T$ trajectory  $\{s^*_0 \dots s^*_{T-1}\} \in D$, we first make the assumption that $s^*_{T-1}$ is a terminal goal state, and so its reward is assigned to $1$. 

Note that the expected value of some state $s_k$ can be expressed as:

\begin{equation}
V(s_k) = \mathbb{E} [\sum_{t=0}^{T-1} \gamma^t r_{t+k}]
\end{equation}
Because the reward at $s^*_{T-1}$ is $1$, we can assign the values using samples from the demonstration as:

$$
V(s^*_{T-1}) = 1, V(s^*_{T-2}) = \gamma, V(s^*_{T-3}) = \gamma^2 \dots, V(s^*_0) = \gamma^{T-1}.
$$
In general, we express the value of some observation $s^*_t$ as:
\begin{equation}
    V(s^*_t) = \gamma^{T-t-1}.
\end{equation}
This update is shown in figure~\ref{fig:value_assignment}. Here $T-t-1$ is effectively the number of steps remaining in the trajectory. It corresponds to how much the value at the goal will be discounted from state $t$ before reaching the terminal state. Because we have a sequence of optimal expert observations, we~\emph{know} how many steps remain. 

We use a deep neural network to learn the values, and aim to minimize the following loss:

\begin{equation}
L_\theta = \Vert V_\theta(s_t) - \gamma^{T-t-1}\Vert^2.
\end{equation}
This simple yet effective approach is shown in Algorithm~\ref{algo:pvo}.

\subsection{Step 2: Learning action-values from values}
\begin{figure}[htb]
    \centering
        \includegraphics[width=.35\linewidth]{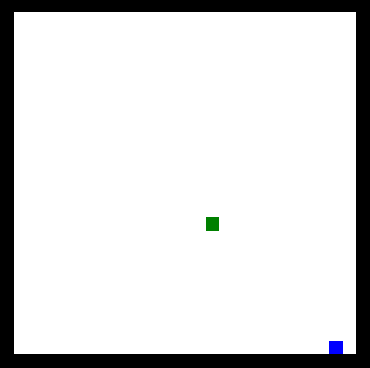}
        \includegraphics[width=.35\linewidth]{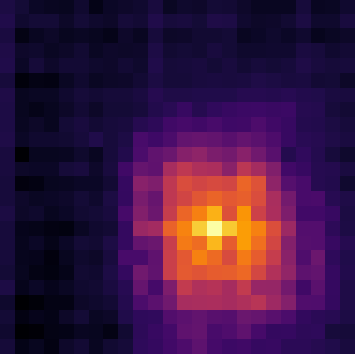}
    
        \includegraphics[width=.35\linewidth]{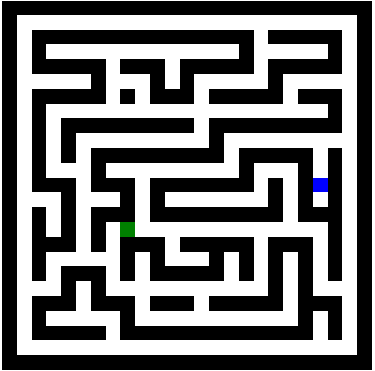}
        \includegraphics[width=.35\linewidth]{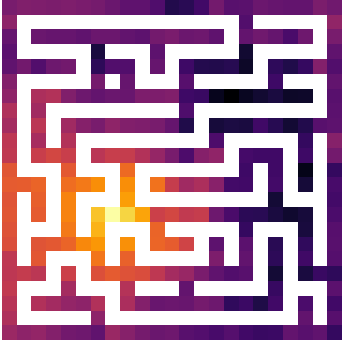}
    \caption{Heatmap of values learned by PVO in unseen maze environments. Brighter colors have larger values.}
    \label{fig:ivo_values}
\end{figure}
\begin{figure}[htb]
    \centering
        \includegraphics[width=.185\linewidth]{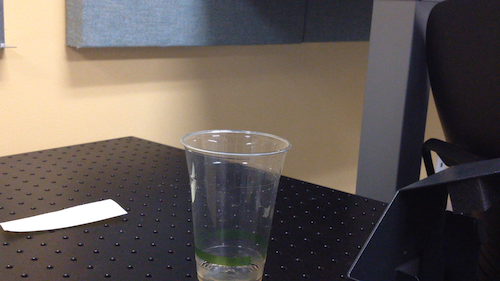}
        \includegraphics[width=.185\linewidth]{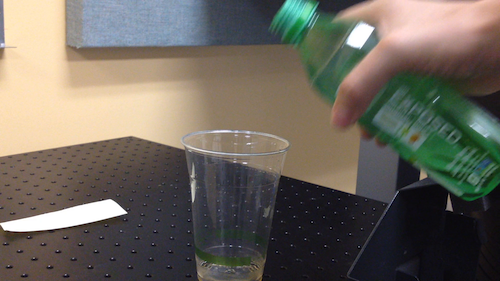}
        \includegraphics[width=.185\linewidth]{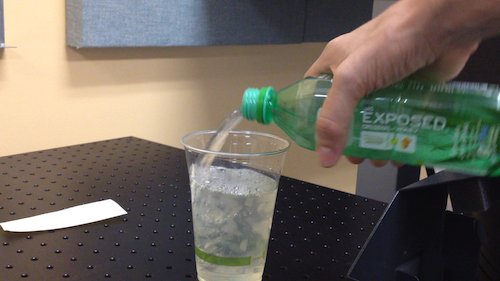}
        \includegraphics[width=.185\linewidth]{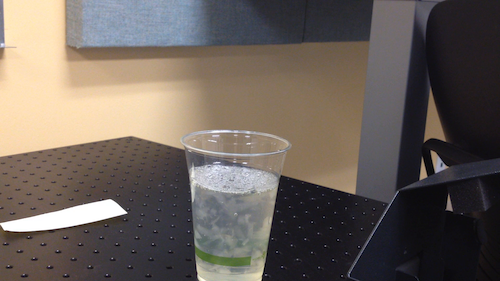}
    
        \includegraphics[width=.95\linewidth]{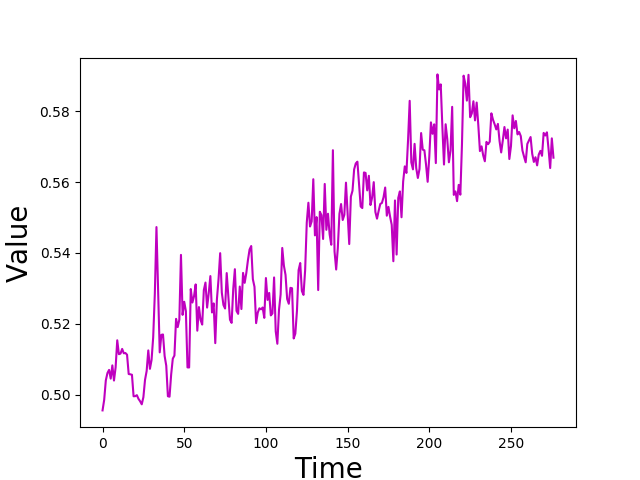}
    \caption{Values learned by PVO in the pouring dataset. The top row represents select frames from a single, unseen video. The bottom row represents learned values for each frame in the video.}
    \label{fig:pour_values}
\end{figure}

Given the learned values, we aim to use RL to learn action-values and a corresponding policy. We introduce two approaches to this problem: 1) using the values to replace bootstrapping in Q-learning and 2) using the values as a potential-based shaping reward. 

\subsubsection{Replacing bootstrapping in Q-learning}
The typical loss update for Q-learning can be defined as:

\begin{equation}
L_\theta = \Vert y_i - Q_\theta(s_{t}, a_{t}) \Vert^2,
\label{eq:q_values}
\end{equation}

where $y_i= r + \gamma \max_{a_{t+1}} Q_\theta(s_{t+1}, a_{t+1})$. The problem with this approach is that it requires making estimates based off of a moving target. We aim to remove this bootstrapped step by replacing the target network with our estimate of the value function.

The Bellman equation states that the maximal action-value is equivalent to the value of a state under the optimal policy~\cite{suttonbarto}:

\begin{equation}
\max_a Q^*(s,a) = V^*(s).
\end{equation}

Given this definition, we can replace the max operator from equation~\ref{eq:q_values} with the learned value function $V_\theta$, and modify the target accordingly:

\begin{equation}
y_i = r + \gamma V_\theta(s_{t+1})
\end{equation}

Because we assume a sparse reward obtained only at the goal,  and because we do not compute action-values at terminal states, $r$ can be replaced with the surrogate reward of $0$, and so the target becomes:

\begin{equation}
y_i = \gamma V_\theta(s_{t+1}).
\end{equation}

\subsubsection{Potential-based shaping reward}
If the value function is incorrect for some states, using it as a replacement for bootstrapping might be too strong of a signal. That is because the formulation aims to directly maximize the value function, and so may get stuck in locally sub-optimal areas if it is not truly optimal.

As such, we also introduce using a potential-based shaping reward~\cite{ng1999policy}:

\begin{equation}
r = \gamma V_\theta(s_{t+1}) - V_\theta(s_t).
\end{equation}

\section{Experiments}
\begin{figure}[htb]
    \centering
        \includegraphics[width=.185\linewidth]{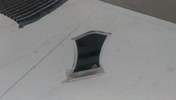}
        \includegraphics[width=.185\linewidth]{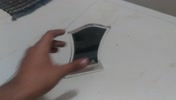}
        \includegraphics[width=.185\linewidth]{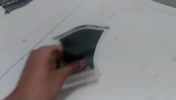}
        \includegraphics[width=.185\linewidth]{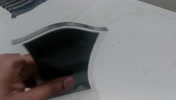}
    
        \includegraphics[width=.95\linewidth]{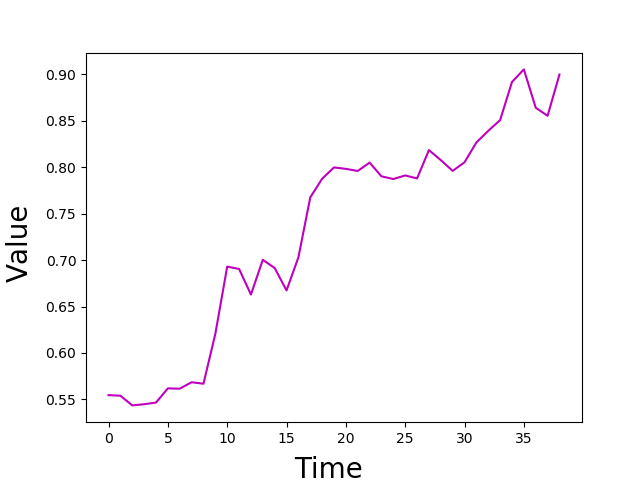}
    \caption{Values learned by PVO in the something something dataset. The top row represents select frames from a single, unseen video. The bottom row represents learned values for each frame in the video.}
    \label{fig:something_values}
\end{figure}

Our experiments aim to demonstrate that PVO can learn values from observation only and that these values can be used to train reinforcement learning agents. We evaluate the agent within unseen environments and aim to determine if PVO learns a general value function that can infer values outside of the training environments. 

\subsubsection{Environments}
In this section, we discuss the environments used for evaluation. We were interested in goal-directed tasks that consisted of a desired target state. We were additionally interested in demonstrating generalization and thus also evaluated within procedurally generated environments.

\subsubsection{Maze environment}
The maze environment, shown in figure~\ref{fig:ivo_values}. consists of procedurally generated mazes. The agent can take actions up, down, left, and right. The game ends when the agent (blue) reaches some target goal (green). We used $A^*$ search to obtain demonstrations in this environment. The demonstration set only consisted of mazes from sizes 4x4 to 20x20. We aim to determine if PVO can learn values in unseen mazes of size 25x25. We obtained 1000 episodes of demonstrations for a simple empty maze and a more complicated one where the agent must navigate around obstacles to reach the goal.

\subsubsection{Liquid pouring dataset}
The liquid pouring dataset has been used to train robots to learn to pour from videos of humans~\cite{sermanet2016unsupervised}. We use $10$ pouring demonstrations to train values and aim to determine if PVO can infer values in an unseen video.

\subsubsection{Something something dataset}
The something something dataset~\cite{goyal2017something} consists of videos of humans doing something to something, for example, pouring something into something, plugging something into something, etc. We use $60$ videos of people picking up something from a surface to determine if PVO can infer values in an unseen video.

\subsubsection{CoinRun environment}
\begin{figure}[htb]
\centering
 \includegraphics[width=.95\linewidth]{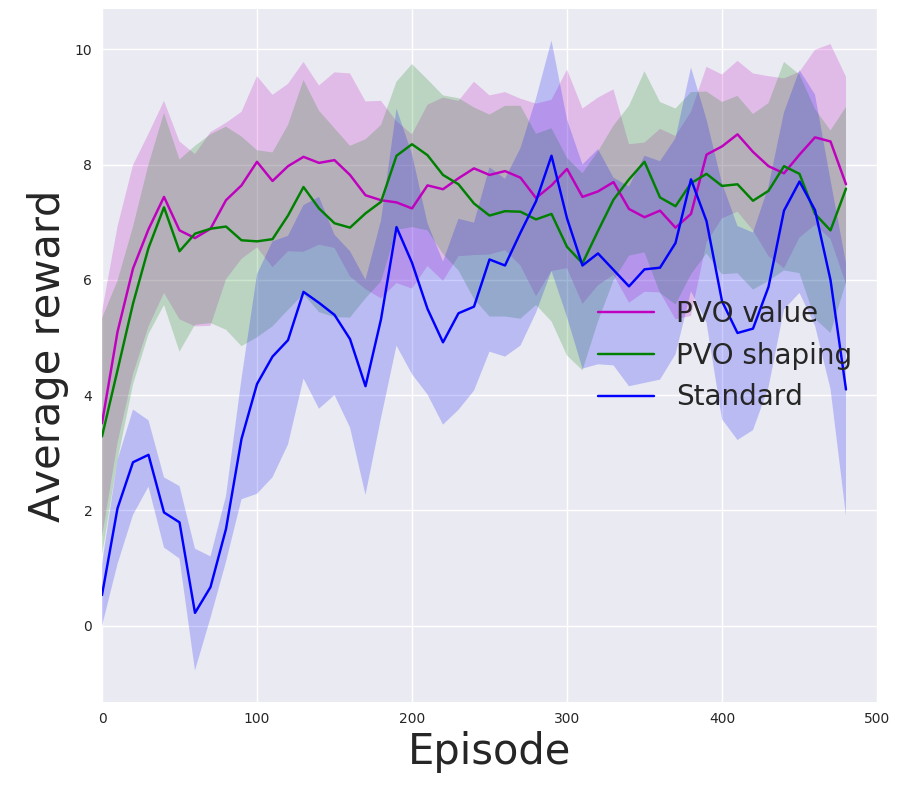}
     \caption{CoinRun reinforcement learning results. The trials were averaged over $5$ runs with a different, unseen, procedurally generated level for each method. The policy was evaluated for $10$ runs every $10$ steps.}
       \label{fig:coin_results}
\end{figure}
The CoinRun environment~\cite{cobbe2018quantifying} consists of procedurally generated platform environments. The background, player, enemies, platforms, obstacles, and goal locations are all randomly instantiated. The agent can take actions left, right, jump, and down, jump-left, jump-right, and do-nothing. The game ends when the agent reaches a single coin in the game. We trained PPO~\cite{schulman2017proximal} for 2.5 million steps to obtain 1000 episodes of expert demonstrations. We evaluate on unseen easy levels.

\subsection{Results} 
In this section, we discuss the results of using PVO to learn values and to train RL agents. Our experiments in the maze environment aim to demonstrate that PVO can learn meaningful values in unseen environments. Figure~\ref{fig:ivo_values} shows a heatmat of the values learned using this approach. It is clear that not only is PVO capable of detecting where the goal is, it can also infer the values of states around the goal.

We also demonstrate value learning in the liquid pouring task, as shown in figure~\ref{fig:pour_values}. PVO has clearly learned a meaningful value function for this task, even though it was only trained with 10 demonstrations. The initial image is an empty glass without any pouring and the value is clearly low. As the glass becomes more full, the values increase. 

Finally, we show value learning for the ``picking up something" task in the something something dataset, as shown in figure~\ref{fig:something_values}. PVO has again learned meaningful values that increase as the task becomes completed.

Our experiments in the CoinRun environment aim to demonstrate that PVO can be used for training reinforcement learning agents in unseen environments. The results are shown in figure~\ref{fig:coin_results}. We call the PVO method that replaces bootstrapping with the learned values~\emph{PVO value} and the method that uses the values as a shaping reward~\emph{PVO shaping}. Both methods learn significantly faster than standard RL. We have thus demonstrated that PVO can be used for imitation and can generalize to unseen environments after receiving observation data only. Additionally, PVO can be used to replace bootstrapping for RL, but the shaping reward was also powerful. One reason for this may be that using the value function to replace bootstrapping essentially initializes the Q-values, which has been shown to be equivalent to potential-based reward shaping~\cite{wiewiora2003potential}.

\section{Conclusion}
In this paper, we have demonstrated that PVO is able to learn values for difficult tasks, and that it can be used to train reinforcement learning agents. We have shown that this approach can generalize to unseen configurations. Finally, we have demonstrated that PVO can significantly speed up reinforcement learning within sparse reward settings.
\bibliographystyle{icml2019}
\bibliography{main}

\end{document}